\documentclass[onecolumn]{article}
 \usepackage[usenames,dvipsnames,svgnames]{xcolor}
 \usepackage{geometry}

\usepackage{natbib}
\usepackage{url}
\usepackage{xspace} 
\usepackage{amsmath}
\usepackage[separate-uncertainty=true,binary-units=true]{siunitx}
\usepackage{mhchem}
\usepackage{array,multirow}

\DeclareMathOperator{\Tr}{\ensuremath{Tr}} 

\usepackage{amsmath}
\def\R{\mathbb{R}}
\usepackage{amssymb}
\usepackage[figuresright]{rotating}
\usepackage[ruled,vlined]{algorithm2e}
\usepackage{ulem} 
\usepackage{xspace} 
 \usepackage[usenames]{xcolor}
 \usepackage{graphicx}
\usepackage{tikz}
\usepackage{pifont}

\newcommand{\gcgc}{GC$\times$GC\xspace}
\newcommand{\gcgcms}{GC$\times$GC$-$MS\xspace}
\newcommand{\gcgcfid}{GC$\times$GC$-$FID\xspace}
\newcommand{\barc}{BARCHAN\xspace}
\newcommand{\cfit}{Curfit2D\xspace}
\newcommand{\tco}{\ding{172}\xspace}
\newcommand{\tct}{\ding{173}\xspace}

\usetikzlibrary{matrix,calc,shapes,positioning}

\tikzset{
	boxNode/.style = {draw,rectangle,rounded corners=3pt,node distance=2cm},
}

\title{\barc: Blob Alignment for \\Robust CHromatographic ANalysis}

\author{Camille Couprie, Laurent Duval, Maxime Moreaud,\\ Sophie H\'enon, M\'elinda Tebib, Vincent Souchon}

\begin{document}
\maketitle

\begin{abstract}
Two dimensional gas chromatography (\gcgc) plays a central role into the elucidation of complex samples. The automation of the identification of peak areas is of prime interest to obtain a fast and repeatable analysis of  chromatograms. To determine the concentration of compounds or pseudo-compounds, templates of blobs are defined and superimposed on a reference chromatogram. The templates then need to be modified when different chromatograms are recorded. In this study, we present a chromatogram and template alignment method based on peak registration called \barc. Peaks are identified using a robust  mathematical morphology tool. The alignment is performed by a probabilistic estimation of a rigid transformation along the first dimension, and a non-rigid transformation in the second dimension, taking into account noise, outliers and missing peaks in a fully automated way. Resulting aligned chromatograms and masks are presented on two datasets. The proposed algorithm proves to be fast and reliable. It significantly reduces the time to results for \gcgc analysis.
\end{abstract}

\begin{quote}
Published in  Journal of Chromatography A (J. Chrom. A.), 2017, 1484, pages 65--72,  Virtual Special Issue RIVA 2016 (40th International Symposium on Capillary Chromatography and 13th GCxGC Symposium)

 \url{http://dx.doi.org/10.1016/j.chroma.2017.01.003}
\end{quote}

\begin{quote}
\end{quote}

\section{Introduction}

First introduced in 1991 by Phillips \emph{et al.} \cite{Liu_Z_1991_j-chromatogr-sci_comprehensive_tdgcuoctmi}, comprehensive two-dimensional gas chromatography (\gcgc) has become in the past decade a highly popular and powerful analytical technique for the characterization of many complex samples such as food derivatives, fragrances, essential oils or petrochemical products \cite{Adahchour_M_2008_j-chrom-a_recent_dactdgc,Meinert_C_2012_j-angew-chem-int-ed_new_dssctdgc,Seeley_J_2012_j-chrom-a_recent_afcmgc,Cortes_H_2009_j-sep-sci_comprehensive_tdgcr}. In the field of oil industry, \gcgc gives an unprecedented level of information \cite{Vendeuvre_C_2005_j-chrom-a_characterization_mdctdgcgcgcpapvsamd} thanks to the use of two complementary separations combining different selectivities. It is very useful in the understanding of catalytic reactions or in the design of refining process units \cite{Bertoncini_F_2013_book_gas_c2dgcpirs,Nizio_K_2012_j-chrom-a_comprehensive_msap}.

From an instrumental point of view, much progress has been made since the early nineties on both hardware and modulation systems \cite{Edwards_M_2011_j-anal-bioanal-chem_modulation_ctdgc20yi}. Many modulator configurations are depicted in the literature or nowadays sold by manufacturers. With the use of leak-free column unions, many of these systems have become robust, easy to use, without cryogenic fluids handling while providing high resolution. Within a series of several consecutive injections, almost no significant shifts in retention times are observed and repeatability of \gcgc experiments is nowadays a minor problem. However, reproducibility of \gcgc results for detailed group-type analysis on complex mixtures is still a great challenge due to column aging, trimming or to slight differences in column features. This results in shifts on retention times that can affect the proper quantification of a single compound, a group of isomers or pseudo-compounds. Experimental retention time locking (RTL) procedures have been proposed to counterbalance shifts on retention times but these procedures must be repeated regularly \cite{Mommers_J_2011_j-chrom-a_retention_tlpctdgc}. On the way to routine analysis for \gcgc, data treatment has therefore become the preferred option  to reduce the time to results \cite{Vendeuvre_C_2007_j-ogst_comprehensive_tdgcdcpp,Murray_J_2012_j-chrom-a_qualitative_qactdgc,Reichenbach_S_2012_j-chrom-a_features_ntcsactdc,Zeng_Z_2014_j-trac-trends-anal-chem_interpretation_ctdgcduac}. 

A common way of treating \gcgc data is to quantify compounds according to their number of carbon atoms and their chemical families by dividing the 2D chromatographic space into contiguous regions that are associated to a group of isomers. This treatment benefits from the group-type structure of the chromatograms and from the roof-tile effect for a set of positional isomers. For example,  for a classical diesel fuel, up to \num{300} or \num{400} regions
 (often referred to as blobs)  may be defined. Due to the lack of robustness in retention times, this step often requires human input and is highly time-consuming when moving from an instrument to another or when columns are getting degraded. Several hours may be necessary to correctly recalibrate a template of a few hundreds of blobs on a known sample. This operator-dependent step causes variability in  quantitative results which is detrimental to reproducibility. In that goal, 2D-chromatogram alignment methods, consisting in modifying a recently acquired chromatogram to  match  a reference one, have been a quite active research area.

In this paper, we propose a new algorithm called \barc\footnote{The name is inspired by wind-produced crescent-shaped sand dunes (\emph{barkhan} or \emph{barchan}) reminiscent of 2D chromatogram shapes.}  which aims at aligning \gcgc chromatograms. It relies on a first peak selection step and then considers the alignment of the two point sets as a probability density estimation problem. This algorithm does not require the placement of anchor points by the user.

\section{Material and methods}

\subsection{Datasets and \gcgc methods}
The straight-run gas-oil sample named GO-SR which is used in this study was provided by IFP Energies nouvelles and was analyzed on different experimental set-ups.  Its boiling point distribution ranges from \SIrange{180}{400}{\celsius}.

Dataset 1 was built by considering two \gcgc chromatograms obtained on two different experimental set-ups in the same operating conditions with cryogenic modulation. These \gcgc experiments were carried out with an Agilent 7890A chromatograph (Santa Clara, California, USA)  equipped with a split/splitless injector, a LN2 two-stage 4 jets cryogenic modulation system from LECO (Saint-Joseph, Michigan, USA) and an FID. The two evaluated column sets were composed of a first 1D apolar 1D HP-PONA column (\SI{20}{\meter}, \SI{0.2}{\milli\meter}, \SI{0.5}{\micro\meter}, J\&W, Folson, USA) and a mid-polar BPX-50 column (\SI{1}{\meter}, \SI{0.1}{\milli\meter}, \SI{0.1}{\micro\meter}, SGE, Milton Keynes, United Kingdom) connected together with Siltite microunions from SGE. Experiments were run with a constant flow rate of \SI{1}{\milli\liter\per\minute}, a temperature program from \SI{60}{\celsius} (\SI{0.5}{\minute}) to \SI{350}{\celsius} at \SI{2}{\celsius\per\minute}, a \SI[retain-explicit-plus=true]{+30}{\celsius} offset for hot jets and a \SI{8}{\second} modulation period. \SI{0.5}{\micro\liter}   of neat sample were injected with a \num{1/100} split ratio. 

Dataset 2 includes a reference chromatogram obtained in the previous conditions and a chromatogram of the same sample obtained with a microfluidic modulation system. These \gcgc data were obtained on a Agilent 7890B chromatograph equipped a split/splitless injector, a Griffith \cite{Griffith_J_2012_j-chrom-a_reversed-flow_dfmctdgc}  type modulation system supplied by the Research Institute for Chromatography (Kortrijk, Belgium) and a FID. The modulation system consists in two Agilent CFT plates (a purged three-way and a two-way splitter) connected to an accumulation capillary. Separation was performed on a DB-1 (\SI{20}{\meter}, \SI{0.1}{\milli\meter}, \SI{0.4}{\micro\meter}) 1D column and a DB-17HT (\SI{10}{\meter}, \SI{0.32}{\milli\meter}, \SI{0.15}{\micro\meter}, J\&W) 2D column. The modulation period was set to \SI{10}{\second} whereas the oven programming and injection conditions were similar to the ones previously described.

\subsection{Software} 

\barc is implemented in C and Matlab. 
In-house platform INDIGO runs it through a user-friendly interface while the proprietary 2DChrom\textregistered\xspace software creates template masks (.idn files) and 2D images from \gcgc data.

\subsection{Calculations} 
The quality of the alignments obtained with \barc was evaluated by two different ways. The correlation coefficient CC \cite{DeBoer_W_2014_j-chrom-a_two-dimensional_spac} as well as  the Structural Similarity index SSIM \cite{Wang_Z_2004_j-ieee-tip_image_qaevss} between the reference chromatogram and the other one were computed. They directly match global image intensities, without feature analysis. Calculation details for CC and SSIM are provided in the supplementary material. These results were obtained on a restricted area of interest defined by the user. A second indicator to evaluate the quality of the alignment is the match quality between the \barc adjusted template and a fully manually registered template. In practice, this featural similarity index is performed by comparing  quantitative results obtained on chemical families with the template mask and with the \barc optimized mask.

\section{Theory}
\subsection{Related works}

We may distinguish two classes of alignment methods: the ones that are directly performed on the full chromatographic signal, and the others which require a prior peak selection step. 
In the first class, the works of \cite{VanMispelaar_V_2003_j-chrom-a_quantitative_atcctdgc} and \cite{Pierce_K_2005_j-anal-chem_comprehensive_tdrtaaecactdsd} look for shifts minimizing a correlation score between  signals. In \cite{Hollingsworth_B_2006_j-chrom-a_comparative_vctdgc}, an affine transformation is assumed between the two chromatograms to register. The recent work of de Boer \cite{DeBoer_W_2014_j-chrom-a_two-dimensional_spac} looks for a warping function parametrized with splines that transforms the chromatogram to be registered into a chromatogram aligned with the reference. Low-degree polynomial alignment is proposed in \cite{Reichenbach_S_2015_j-anal-chem_alignment_ctdgcdscd}. Full image registration \cite{Zitova_B_2003_j-image-vis-comput_image_rms} is however limited for applications in \gcgc because of the variability in chromatograms: positions of peaks in the two chromatograms could be similar, but this is not the case of their intensities. Therefore, the majority of alignment methods choose to first extract peaks in the reference and target chromatograms to only register the informative parts of chromatographic images.

Thus, among  approaches dedicated to chromatogram alignment, the work of \cite{VanMispelaar_V_2005_j-chrom-a_classification_hscoudsctdgcmt} (focused on quantitative analysis) deduces local peaks displacements by correlation computations in slightly shifted blocks surrounding peaks. Variations of peak patterns in different experimental conditions (e.g. temperature) is studied in \cite{Ni_M_2005_j-chrom-a_peak_pvrctdgca}, and exhibits satisfactory results for estimating an affine transformation. Similarly, \cite{Reichenbach_S_2009_j-chrom-a_smart_tppmctdlc} also models rigid transformations for LC$\times$LC (2D liquid chromatography) template alignment. However, these hypotheses appear to be too restrictive in a general setting. Therefore \cite{Zhang_D_2008_j-anal-chem_two-dimensional_cowaagcgcmsd} extended the space of possible deformations by looking for a warping function that transforms signals.  Correlation Optimized Warping (COW) is judged effective by \cite{VanNederkassel_A_2006_j-chrom-a_comparison_taca} that compares three different registration approaches, including target peak alignment (TPA) and semi-parametric time warping (STW) for one specific analysis. However, COW is still not satisfactory when incomplete separation and co-elution problems exist as pointed out by \cite{Parastar_H_2012_j-chem-int-lab-syst_comprehensive_tdgcgcgcrtscmubpacosmcr}. Instead, the latter  uses bilinear peak alignment in addition to COW to correct for progressive within run retention time shifts on the second chromatographic dimension. In \cite{Weusten_J_2012_j-anal-chim-acta_alignment_csgcgcmsfucm}, the alignment is performed after embedding the chromatograms surfaces into a three-dimensional cylinder, and the parametrization of the transform employs polynomials. The DIstance and Spectrum Correlation Optimization (DISCO) alignment method of \cite{Wang_B_2010_j-anal-chem_disco_dscoatdgctofmsbm}, extended in \cite{Wang_B_2012_incoll_disco2_cpaatdgctofms}, uses an elaborate peak selection procedure followed by interpolation to perform the alignment.  The approach from \cite{Kim_S_2011_j-bmc-bioinformatics_smith-waterman_pactdgcms} also performs peak alignment via correlation score minimization using dynamic programming, comparing favorably to DISCO. Finally, the work of \cite{Gros_J_2012_j-anal-chem_robust_aatdc}  performs an assessment of different \gcgc alignment methods with a new one. Their method requires a manual placement of matching peaks pairs, then the registration is performed differently on each axis: linear deformations along one dimension, and a neighbor based interpolation in a Voronoi diagram defined using the alignment anchor points for the other dimension. The linear constraint  is relevant because one dimension displacements are independent of the other dimension elution times. The requirement of user-defined alignment points is robust to large variations in the reference and target chromatogram, at the expense of time-consuming markers placement.

\begin{figure}[htb]
\caption{Flowchart methodology for \barc.\label{fig:flowchart}}
\vspace{0.2cm}
\centering
\begin{tikzpicture}[thick,scale=1, every node/.style={transform shape}]
\node[boxNode](n11) at (0,0) {\shortstack{New \gcgc data\\image file: .png, .bmp, .jpg}};
\node[boxNode,right= of n11.north east,anchor=north west](n12)  {\shortstack{Reference \gcgc chromatogram\\ (image file) with or without \\an associated  reference \\template mask (.idn file)
}};
\node[boxNode](n20)  at (3.35,-2.5) {\shortstack{Definition of an \\area of interest}};
\node[boxNode](n30)  at (3.35,-3.5) {Feature point extraction};
\node[boxNode](n40)  at (3.35,-4.6) {\shortstack{Probabilistic alignment estimation: a Gaussian\\  Mixture Model (GMM) is used to model datasets}};
\node[boxNode](n50)  at (3.35,-6) {\shortstack{Optimization with the \\Expectation-Maximization algorithm}};
\node[boxNode,below=9cm of n11.north west,anchor=south west](n61)  {\shortstack{\tco Distorted reference  \\template mask to be \\applied  on  new data }};
\node[boxNode,below=9cm of n12.north east,anchor=south east](n62)  {\shortstack{\tct Distorted chromatogram \\(\gcgc image) to match  \\with reference 2D image }};
\draw[->,>=latex] 
	(n11.south) -- (n20.north west);
\draw[->,>=latex] 
	(n12.south) -- (n20.north east);
	\draw[] 
	(n20.south) -- (n30.north)
	(n30.south) -- (n40.north)
	(n40.south) -- (n50.north)
;	

\draw[->,>=latex] 
(n50) |- (3.35,-7) -| (3.35,-7) -- (n62);

\draw[->,>=latex] 
(n50) |- (3.35,-7) -| (3.35,-7) -- (n61);

\end{tikzpicture}
\end{figure}
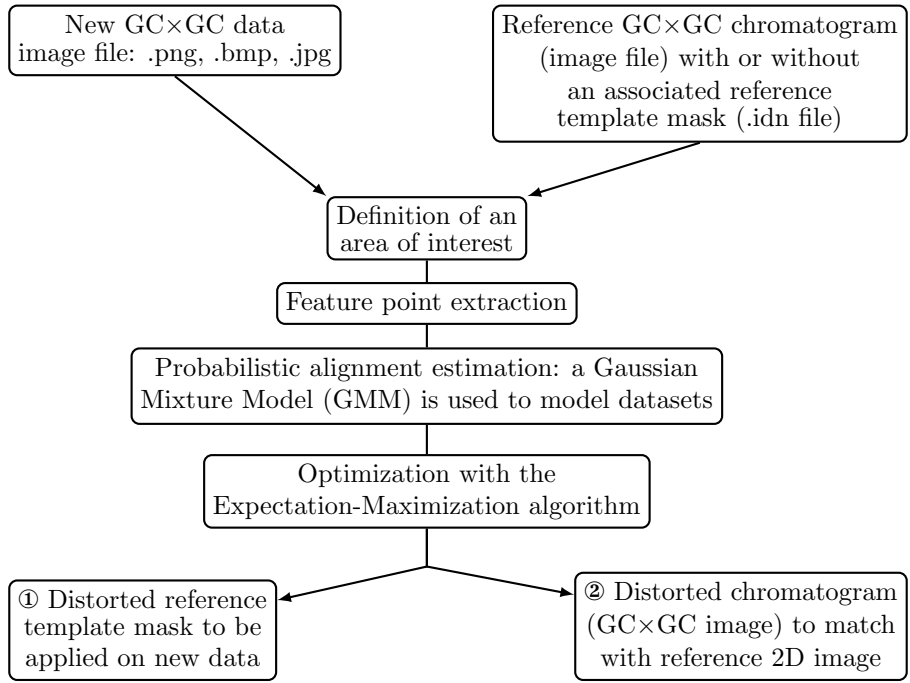

\subsection{\barc methodology}
A schematic view for the principles of \barc is depicted in the flowchart from Figure \ref{fig:flowchart}. 
First, \gcgc chromatograms of the sample to analyze and the reference 2D image are loaded as images files. Then, the user is provided with a brush to surround, in a user interface, the  area of interest on both  reference and  new 2D chromatograms (see Figure \ref{fig:areas}). Peaks are extracted in those areas (Section \ref{sec:hmax}). Only one centroid per local maximum is retained for the point set registration in order to diminish computation times and to prevent bias for large peaks. Datasets are then assimilated to centroids of a Gaussian Mixture Model (GMM) and a weighted noise model is added. Advantage is taken from  recent progresses in point set registration, using a probabilistic and variational approach \cite{Myronenko_A_2010_j-ieee-tpami_point_srcpd}. This choice is motivated by the fact that a complex transformation must be modeled while remaining robust to noise and outliers. In this context, GMMs (Gaussian Mixture Models) are particularly efficient at reconstructing missing data, which is especially convenient when selected peaks in one point cloud are not included in the other one. Finally, model parameters are optimized to yield registered results. Two types of results are produced:

\begin{itemize}
\item if a template mask for the reference chromatogram exists, the transformation of the template points leading to a registered template mask is computed.
\item an aligned chromatogram may also be produced by computing the transformation of a grid defined as the coordinates of every pixel in an image, and interpolating the target image values at the transformed coordinates.
\end{itemize}
Details on the calculations for every step are provided in the next paragraphs.

\subsection{Feature point extraction\label{sec:hmax}}

Despite the good behavior of the employed registration algorithm regarding noise and outliers, it is desirable to extract the most resembling point sets. Therefore, preliminary \gcgc enhancement 
\cite{Ning_X_2014_j-chemometr-intell-lab-syst_chromatogram_bedusbeads,Samanipour_S_2015_j-chrom-a_analyte_qctdgcambcpdmeers}  
proves useful.  

Inherent to the \gcgc experimentation procedure, fragments of the stationary phase are frequently lost by the column resulting into the presence of hyperbolic lines in the chromatogram. Their differentiation from the real peaks is difficult to automate because of possible overlaps with the chromatogram peaks of interest. Therefore, in our treatments, a rough area of interest is delimited by an operator, taking approximately ten seconds.

\begin{figure}[h]
\centering
   \caption{Detection of $h$-maxima: only one peak is detected in this example (dotted line).\label{fig:hmax}}
   \includegraphics[width=0.4\textwidth]{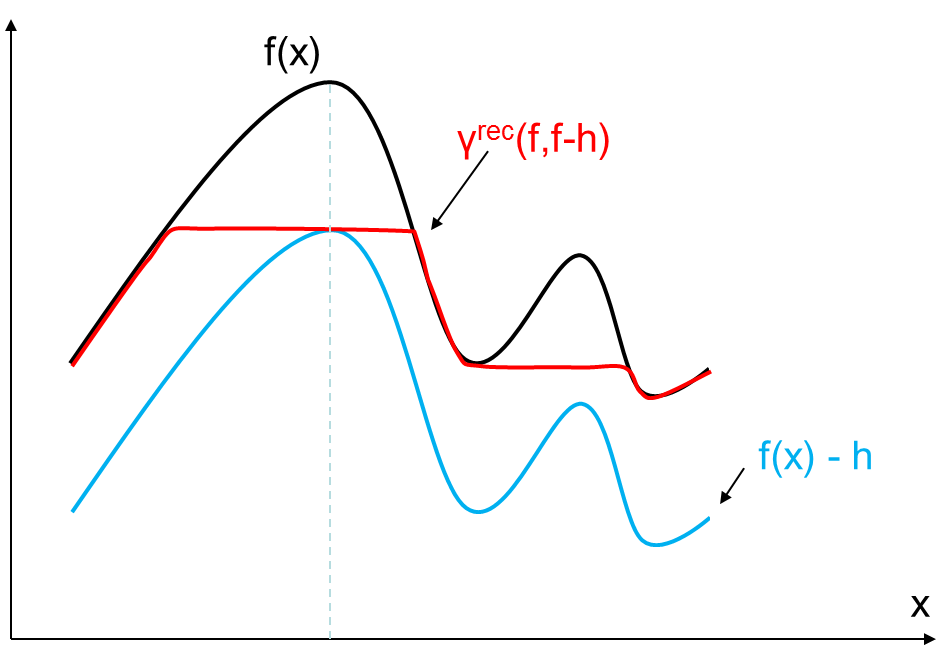}
\end{figure}

Rather than using second or third derivatives of the chromatogram \cite{Fredriksson_M_2009_j-sep-sci_automatic_pfmlcmsdgsdf}, which require non-trivial parameters to set, we employ the approach of \cite[p. 97--106]{Bertoncini_F_2013_book_gas_c2dgcpirs} and extract the $h-$maxima of the chromatograms. 
Simply put, all local maxima having a height greater than a scalar $h$ are extracted.
Starting from an input signal $f$ from $\R^d$ to $\R$, 
the positions of the $h$-maxima may be obtained via a morphological opening by reconstruction, noted $\gamma^{\mbox{rec}}(f,f-h)$. More specifically, this operation is defined as the supremum of all geodesic dilations of $f-h$ by unit balls in $f$. More details are provided in \cite{Bertoncini_F_2013_book_gas_c2dgcpirs} and a scheme is displayed in Figure~\ref{fig:hmax}.

\subsection{Data alignment model}
 
To guarantee results where the first point set is similar to the registered point set, while being robust to noise and outliers, we choose to employ a probabilistic approach.
Supposing that the first point set $X$ follows a normal distribution, the Coherent Point Drift method \cite{Myronenko_A_2010_j-ieee-tpami_point_srcpd} seeks to estimate the probability density that explains the data $X$ as a weighted sum of Gaussians initially centered by the second point set $Y$. 

We introduce our notations as follows. The first point set  of size $N\times2$, corresponding to the coordinates of $N$ peaks  extracted in the target chromatogram, is denoted $X=\{X_1, \ldots, X_N\}$. The second point set $Y=\{Y_1, \ldots, Y_M\}$ of size $M\times2$ corresponds to the peak coordinates in the reference chromatogram and is assimilated to centroids of a GMM. Each component $X_i$ is a vector composed of two coordinates denoted $X^{(1)}_i$ and $X^{(2)}_i$. The vector $X^{(i)}$ denotes the $i^{\scriptsize{\mbox{th}}}$ line of matrix $X$. 
Adding a weighted noise model to the GMM probability density function leads to:
\begin{equation}
p(X_n)= \frac{w}{N} + \sum_{m=1}^{M} \frac{1-w}{2 M \pi \sigma^2} \exp \left( -\frac{\|X_n-T(Y_m)\|^2}{2\sigma^2}\right) 
\end{equation}
where the first term takes into account uniform noise weighted by the parameter $w$ fixed between 0 and 1, $\sigma$ is a variance parameter to estimate, and $T$ is the point cloud transform to estimate.  
In this work, motivated by a failure of global rigid transformation attempts on our data, we modeled two different transforms across the two dimensions. We assume that a rigid displacement occurs along the $y$-axis second very short column, similarly to \cite{Gros_J_2012_j-anal-chem_robust_aatdc}, and non-rigid transformations are allowed on the $x$-axis first normal length column. The underlying assumption is a relative anisotropy of the data: two separate pixels in the vertical direction are distant by a much smaller time interval than those aligned horizontally. The $x$-axis is thus potentially subject to more important nonlinear distortions. Thus, 
we model the transformation $T$ of  point cloud $Y$ as:
\begin{align} 
T(Y^{(1)}) & =sY^{(1)}+t,\\
T(Y^{(2)}) & = Y^{(2)}+G W,
 \end{align}
where $s$ and $t$ are real numbers, respectively a scale and a translation parameter to estimate, and $W$ is a vector of length $M$ of non-rigid displacements to estimate.  
The matrix $G \in R^{M\times M}$ is a symmetric matrix defined element-wise by:

\begin{equation}
G_{ij} = \exp^{- \frac{\|Y_i-Y_j\|}{2\beta} },
\end{equation}  
where $\beta$ is a positive scalar.  
The minimization of the non-negative likelihood leads to the minimization of: 
\begin{equation}
E_1(\sigma, W, s,t)= - \sum_{n=1}^N \log p(X_n).
\end{equation}
A regularization of the weights $W$,  enforcing the motion to be smooth, is necessary for the non-rigid registration,
resulting into the following variational problem:  
\begin{equation}
\min_{\sigma, W, s,t} E =  E_1(\sigma, W, s,t) +\frac{\lambda}{2} \Tr(W^{\top} G W),
\end{equation}
where $\Tr$ denotes the trace operator of a matrix. The estimation of parameters $w$, $\beta$ and $\lambda$ is discussed in generic terms in \cite{Yuille_A_1989_j-ijcv_mathematical_amct} and \cite{Myronenko_A_2010_j-ieee-tpami_point_srcpd}. \barc inherits a similar strategy, within the proposed combined rigid/non-rigid registration procedure. The parameter $w\in [0\,,\,1]$, related to the noise level, is first determined by visual inspection on ten regularly-spaced values. Albeit found to be the most determinant, our chromatograms sharing about the same signal-to-noise ratio, this value is kept constant in all our experiments. For other data  types,  multiple figures illustrating different registrations with varying  amounts of noise and outliers with an appropriate choice of $w$ are presented in  \cite{Myronenko_A_2010_j-ieee-tpami_point_srcpd}. The determination  of the other parameters  and $\lambda$ is also discussed in \cite{Yuille_A_1989_j-ijcv_mathematical_amct}. We have set them to $\beta=2$ and $\lambda=2$ as by default in \cite{Myronenko_A_2010_j-ieee-tpami_point_srcpd}. Slight changes did not affect the registration results sensitively.

\subsection{Optimization}
We employ the Expectation-Maximization (EM) algorithm \cite{Dempster_A_1977_j-r-stat-soc-b-stat-methodol_maximum_lidema} that alternates between: 
\begin{itemize}
\item the E step: we compute the probability $P$ of correspondence for every couple of points. 
\item the M step: we estimate the parameters $\sigma, s, t,$ and $W$. 
To that goal, we compute the partial derivative of $E$ with respect to $\sigma, s, t,$ and $W$ and set them to zero leading to an estimate of every parameter. Details are provided in the supplementary material, as well as the final algorithm itself.   
\end{itemize}

\section{Results and discussion}

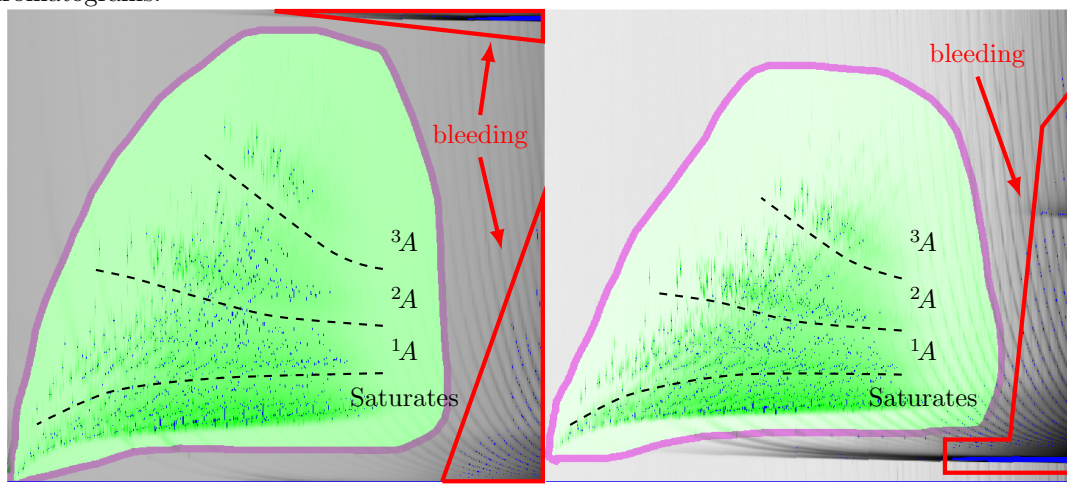
\begin{figure*}[hbt]
\centering
 \caption{\label{fig:areas} Dataset 1:  selection of the areas of interest  (purple lines), new (left) and reference (right) chromatograms.}
\begin{tikzpicture}
\node[anchor=south west,inner sep=0] (image) at (0,0) {\includegraphics[width=0.95\textwidth]{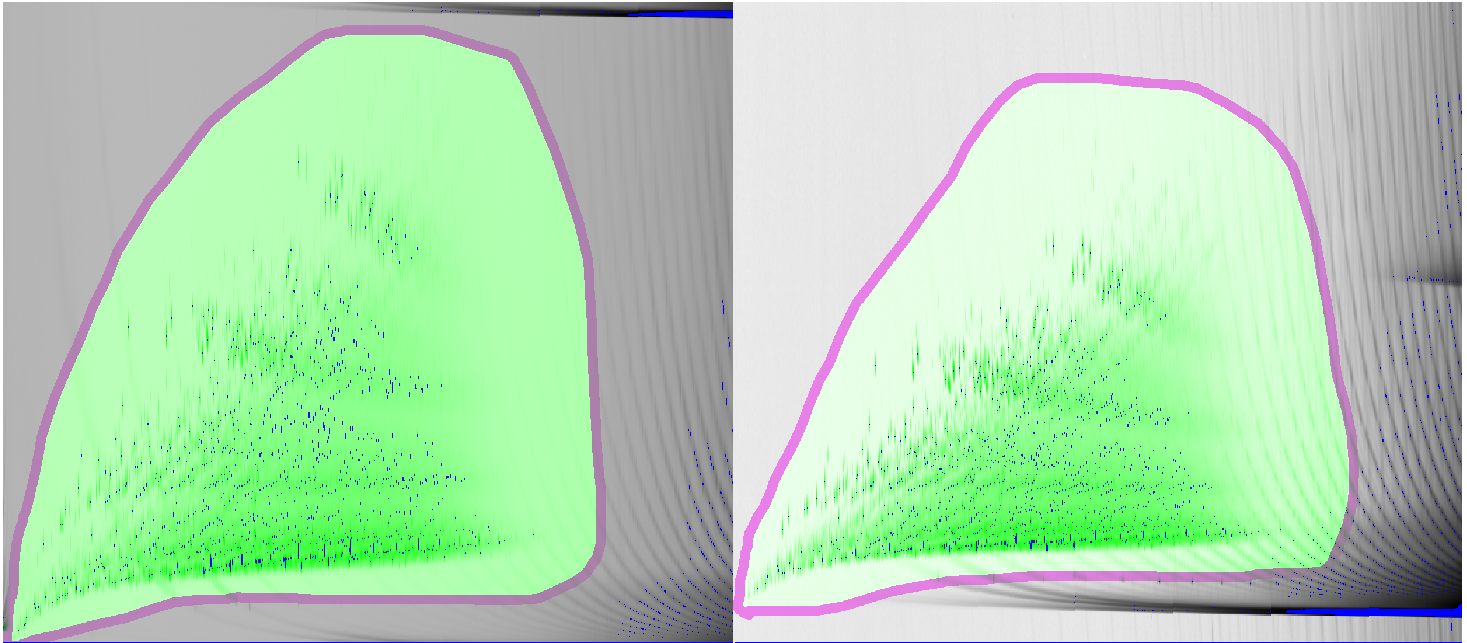}};
\begin{scope}[x={(image.south east)},y={(image.north west)}]
\node[red](n11) at (0.44,0.73) {bleeding};
\draw[red,ultra thick]   (0.405,0.002) -- (0.498,0.002)  -- (0.498,0.6) -- (0.405,0.002) ;
\draw[red,ultra thick]   (0.25,1-0.006) -- (0.498,1-0.006)  -- (0.498,0.93) -- (0.25,1-0.006) ;
\draw[->,>=latex,ultra thick,red] 	(n11.south) -- (0.46,0.49);
\draw[->,>=latex,ultra thick,red] 	(n11.north) -- (0.45,0.93);

\node[red](n12) at (0.9,0.9) {bleeding};
\draw[red,ultra thick]    (1-0.0055,0.02)  -- (1-0.0055,0.85)  -- (0.96,0.75)  -- (0.93,0.09) -- (0.87,0.09) -- 
(0.87,0.02)  -- (1-0.0055,0.02) ;
\draw[->,>=latex,ultra thick,red] 	(n12.south) -- (0.94,0.6);

\node[](a10) at (0.37,0.18) {Saturates};
\node[](a11) at (0.37,0.28) {$^1\!A$};
\node[](a12) at (0.37,0.39) {$^2\!A$};
\node[](a13) at (0.37,0.51) {$^3\!A$};

\node[](a20) at (0.85,0.18) {Saturates};
\node[](a21) at (0.85,0.28) {$^1\!A$};
\node[](a22) at (0.85,0.39) {$^2\!A$};
\node[](a23) at (0.85,0.51) {$^3\!A$};

\draw[thick,dashed] plot [smooth] coordinates {(0.35,0.2+0.030) (0.2,0.18+0.04)  (0.1,0.16+0.032) (0.03,0.09+0.032)};
\draw[thick,dashed] plot [smooth] coordinates {(0.35,0.33) (0.25,0.35)   (0.18,0.39)  (0.12,0.43)  (0.08,0.45) };
\draw[thick,dashed] plot [smooth] coordinates {(0.35,0.45) (0.30,0.49)   (0.18,0.7)  };

\draw[thick,dashed] plot [smooth] coordinates {(0.35+0.48,0.2+0.027) (0.2+0.48,0.2+0.027)  (0.1+0.48,0.16+0.026) (0.52,0.09+0.026)};
\draw[thick,dashed] plot [smooth] coordinates {(0.35+0.48,0.32) (0.25+0.48,0.34)   (0.18+0.48,0.38)   (0.6,0.4) };
\draw[thick,dashed] plot [smooth] coordinates {(0.35+0.48,0.43) (0.30+0.48,0.47)   (0.7,0.6)  };
    \end{scope}
\end{tikzpicture}
\end{figure*}

The areas of interest for both dataset 1 and 2 were defined so that every compound present in the sample is taken into account while limiting the number of peaks due to the bleeding (Figure \ref{fig:areas}). The detected peaks appear as small blue dots on both chromatograms whereas the selected areas are colored in green and delimited with a purple line. Peaks were extracted with a height parameter $h$ from Section \ref{sec:hmax} equal to 120 and 60 for dataset 1 and 2, respectively.

Three types of transformations were evaluated: rigid transformations on both the $x$- and the $y$-axis, non-rigid transformations on both axes and \barc transformation (non-rigid transformation on $x$-axis, rigid on $y$-axis).  They are compared with the \cfit algorithm \cite{DeBoer_W_2014_j-chrom-a_two-dimensional_spac}. Significant changes in  scores, especially for the  CC index, 
 suggest  a better alignment of  two chromatograms for dataset 1 with  \barc. However, small variations in these global indices demonstrate the need for a closer inspection of the results.

\begin{table*}
\centering
\caption{Correlation coefficient (CC) and Similarity index (SSIM) on the two datasets.\label{tab:1}} 
\begin{tabular}{|c|c|c|c|c|c|c|}
  \hline
 & & No registr. & Rigid \cite{Myronenko_A_2010_j-ieee-tpami_point_srcpd} & Non-rigid \cite{Myronenko_A_2010_j-ieee-tpami_point_srcpd} & \cfit \cite{DeBoer_W_2014_j-chrom-a_two-dimensional_spac} & \barc \\
  \hline
\multirow{2}{*}{Dataset 1}  & CC   & \num{0.57} & \num{0.81} & \num{0.94} & \num{0.47} & \num{0.93} \\
 \cline{2-7} 
            & SSIM & \num{0.86} & \num{0.88} & \num{0.95} & \num{0.87} & \num{0.94} \\
\hline
\multirow{2}{*}{Dataset 2} 
 & CC  & \num{0.30} & \num{0.91} & \num{0.81} & \num{0.73}   & \num{0.83} \\
 \cline{2-7} 
						& SSIM & \num{0.91} & \num{0.93} & \num{0.93} & \num{0.95}  & \num{0.94} \\
	\hline
\end{tabular}
\end{table*}

\begin{figure*}[htb]
\scriptsize
\centering
   \caption{\label{fig:point-set} Point sets before and after three different transformations:  rigid, non-rigid and \barc.}
	\begin{tabular}{cc}		
   \includegraphics[width=0.5\textwidth]{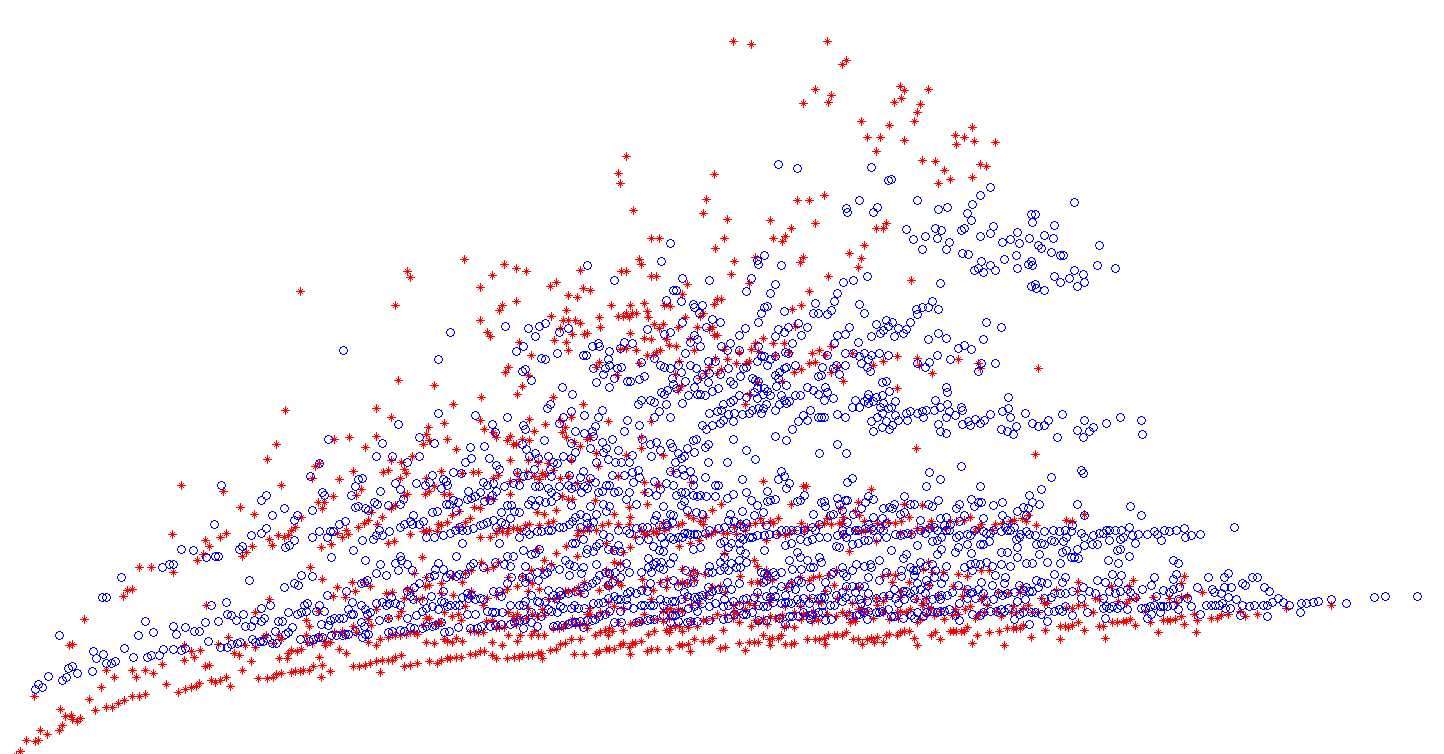}&
	\includegraphics[width=0.5\textwidth]{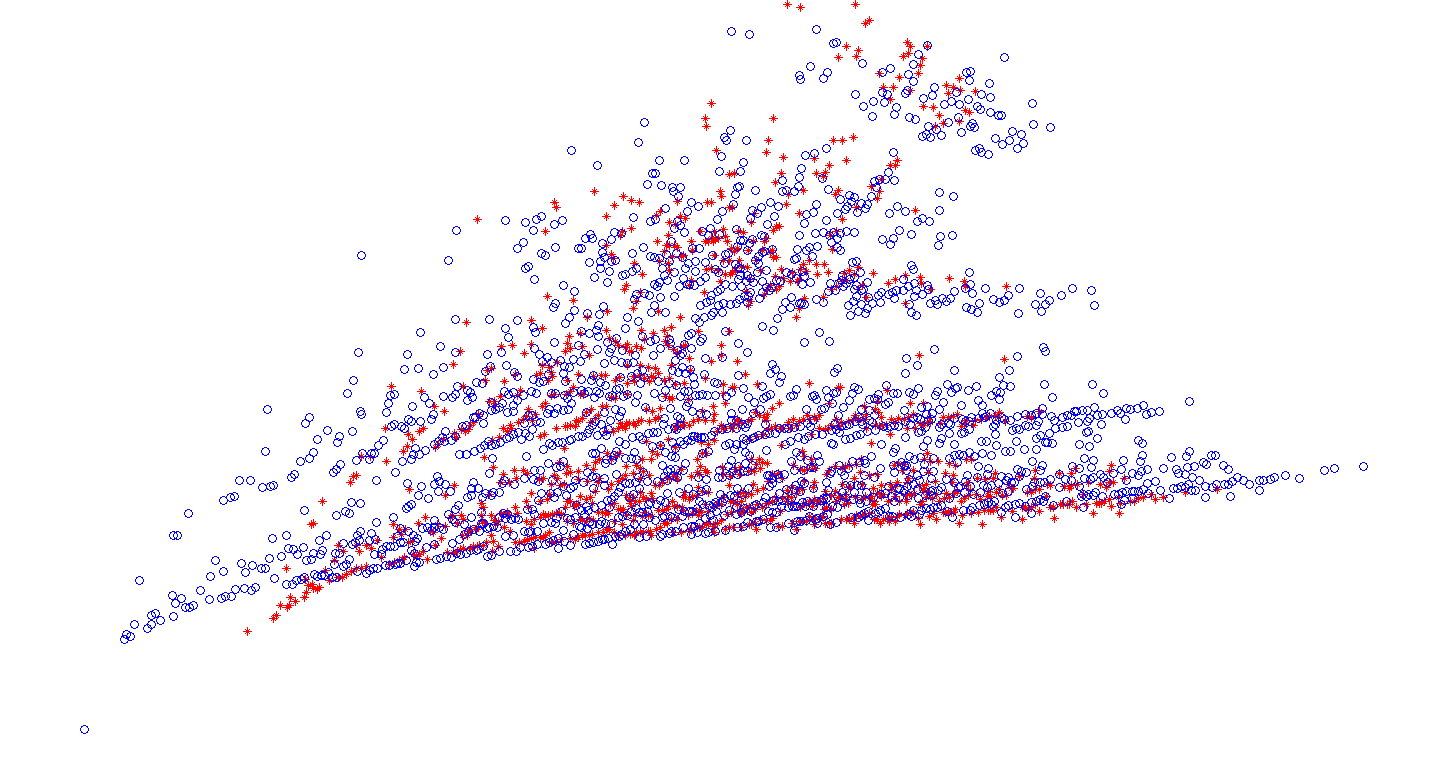}\\
		No transformation (centered cloud points) & Rigid\\
\\
   \includegraphics[width=0.5\textwidth]{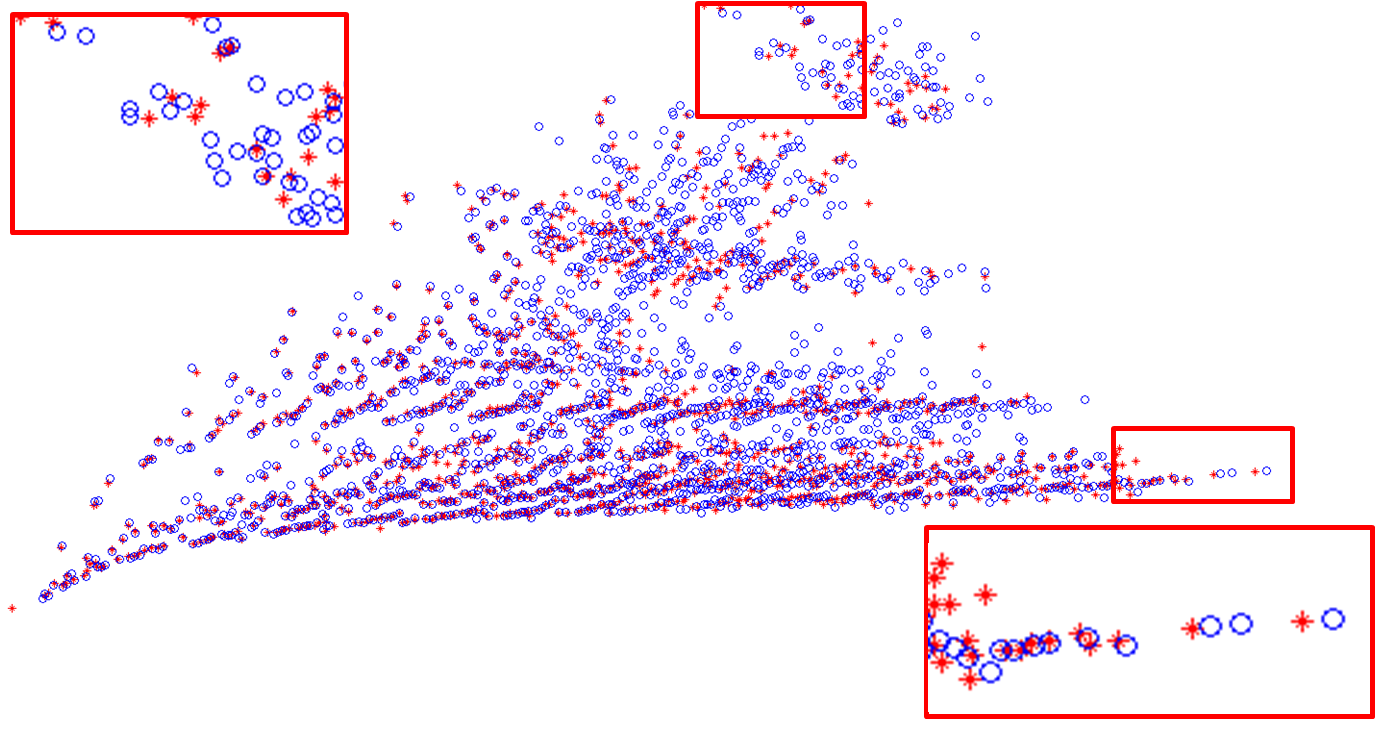}&
	\includegraphics[width=0.5\textwidth]{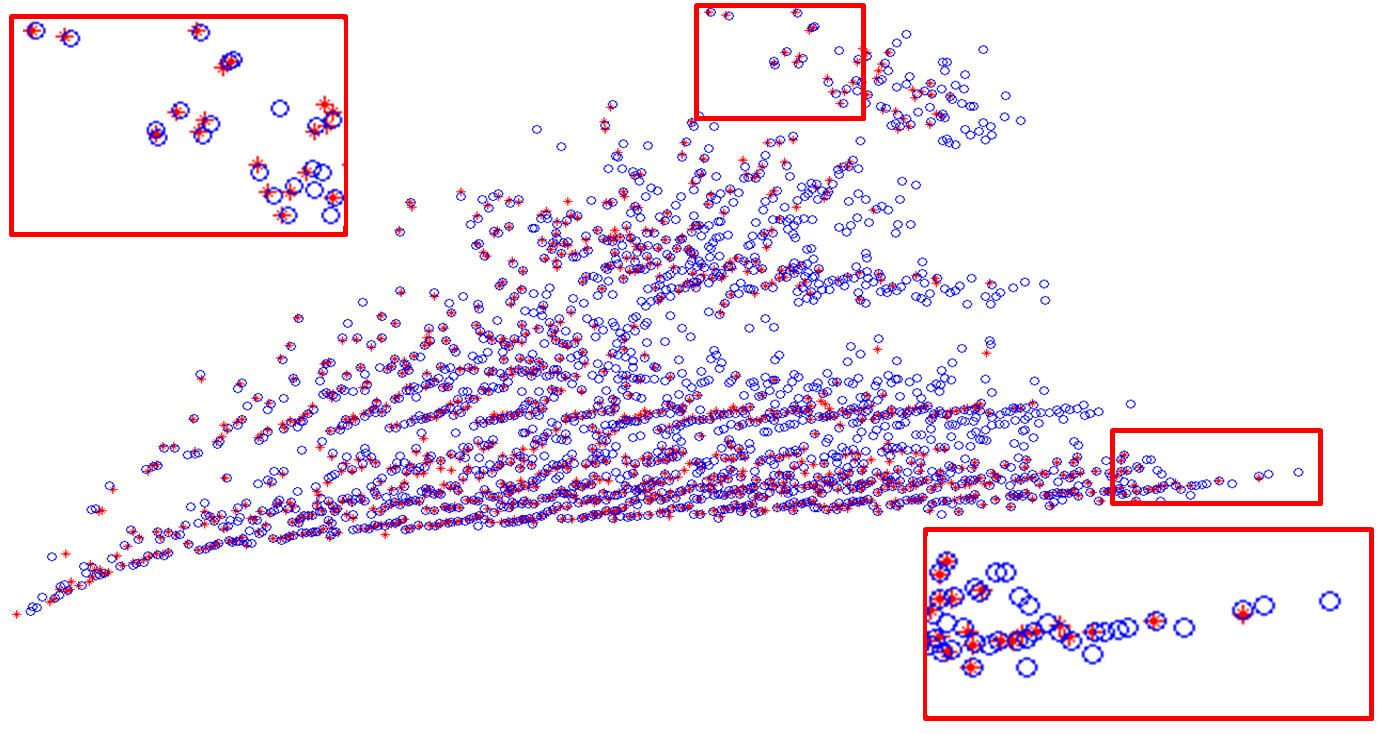} \\
			Non-rigid & \barc
	\end{tabular}
\end{figure*}

Figure \ref{fig:point-set} shows the optimization results for the three tested transformations on dataset 1 thanks to scatterplots \cite{Anscombe_F_1973_j-american-statistician_graphs_sa}. Blue circles correspond to extracted peaks from the reference chromatogram whereas red crosses represent the extracted and transformed peaks for the new \gcgc chromatogram. These images show that a fully rigid transformation (Figure \ref{fig:point-set}, top-right) does not allow a good match between the reference chromatogram and the new one. A better agreement is obtained with the \barc algorithm and the fully non-rigid transformation. However, when looking into details in some specific areas of the 2D chromatogram where the number of extracted peaks is highly different between the reference and the new image (see red boxes at the bottom of Figure \ref{fig:point-set}), \barc algorithm outperforms the fully non-rigid approach. The interest of \barc over the fully non-rigid approach is also shown on the transformation of template masks (see supplementary material). Whereas \barc algorithm leads to a coherent transformation of the template mask including for blobs in the upper right part of the chromatogram which are extrapolated, the fully non-rigid deformation is not relevant.

\begin{figure*}[htb]
\scriptsize
\centering
   \caption{\label{fig:defo} Calculated deformation of the 2D chromatogram with the \barc algorithm on dataset 1.}
	\begin{tabular}{ccc}
   \includegraphics[width=0.31\textwidth]{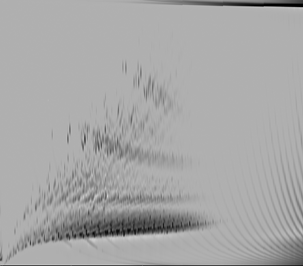}&
   \includegraphics[width=0.31\textwidth]{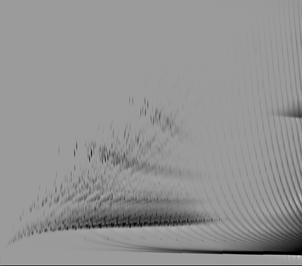}&
	\includegraphics[width=0.31\textwidth]{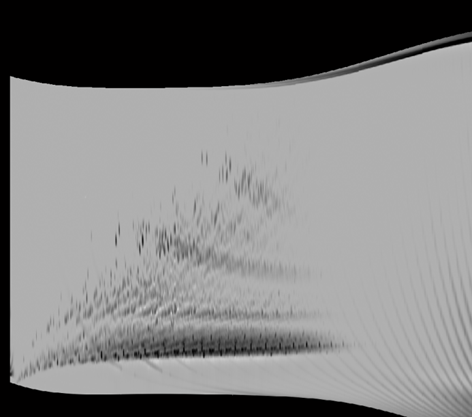}\\
			New. & Reference. & \barc transformed.
	\end{tabular}
\end{figure*}

To illustrate the changes modeled by \barc on the chromatograms, the reference and the new chromatograms from dataset 1 are displayed in Figure \ref{fig:defo}, as well as the resulting aligned chromatogram.

\begin{figure*}[htb]
\scriptsize
\centering
   \caption{\label{fig:reference} Chromatograms with overlaid masks. 	Top: reference and    datasets 1 and 2 transformed with \barc. Middle and bottom: zooms on the top (diaromatics) and bottom (paraffins and naphthenes) red areas.}
	\begin{tabular}{ccc}
   \includegraphics[width=0.33\textwidth]{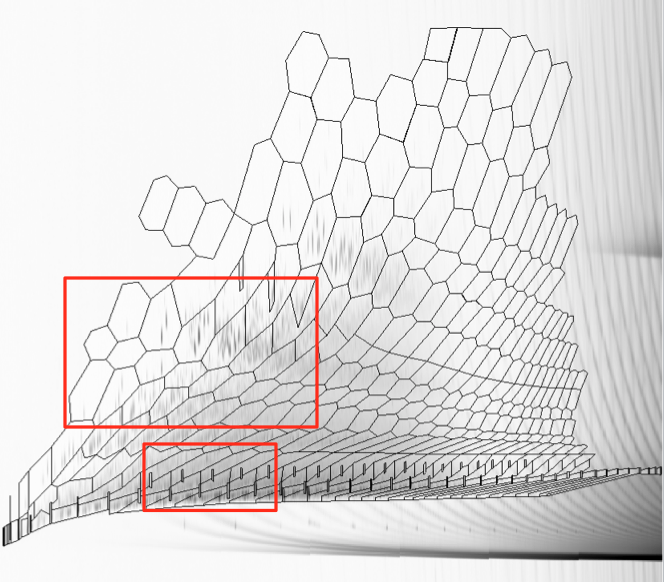}& 
   \includegraphics[width=0.33\textwidth]{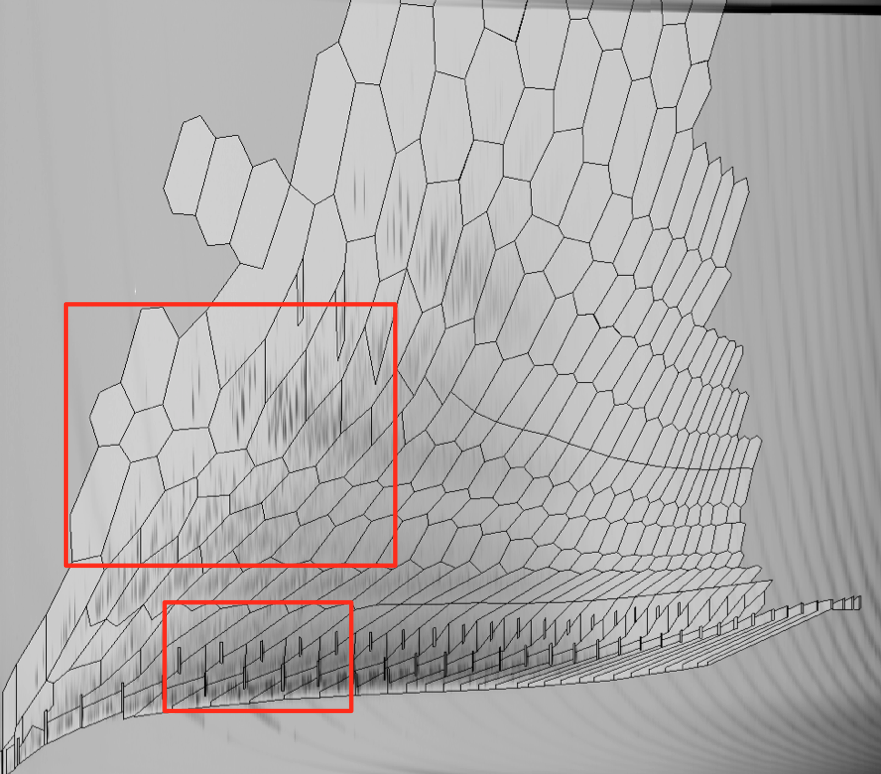}& 
   \includegraphics[width=0.33\textwidth]{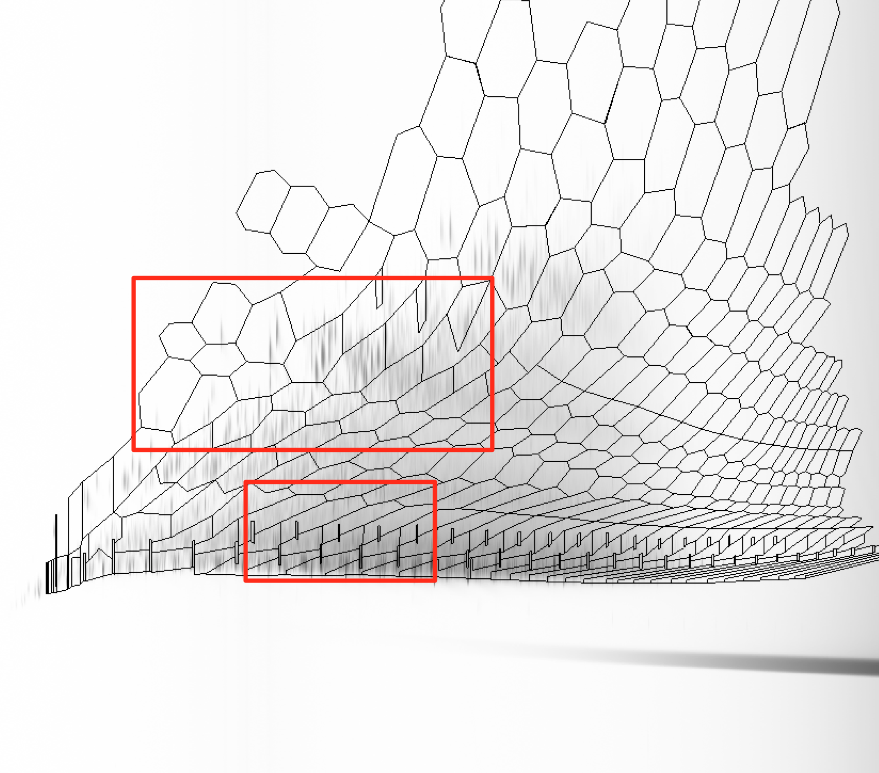}\\ 
    \includegraphics[width=0.33\textwidth, height=0.2\textwidth ]{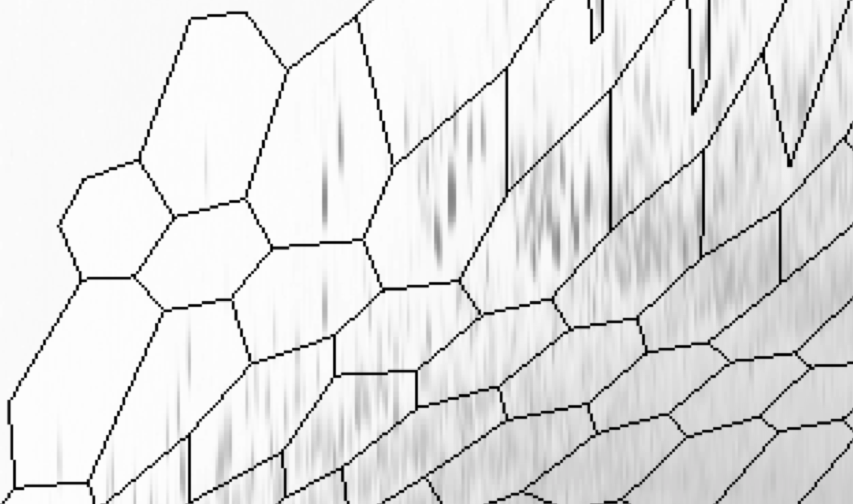}& 
   \includegraphics[width=0.33\textwidth, height=0.2\textwidth]{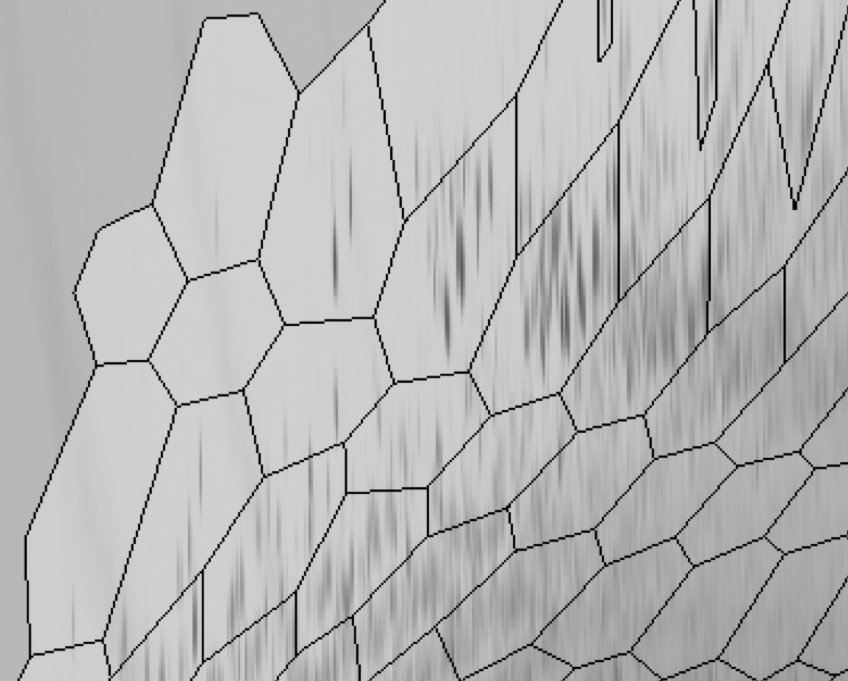}& 
   \includegraphics[width=0.33\textwidth, height=0.2\textwidth]{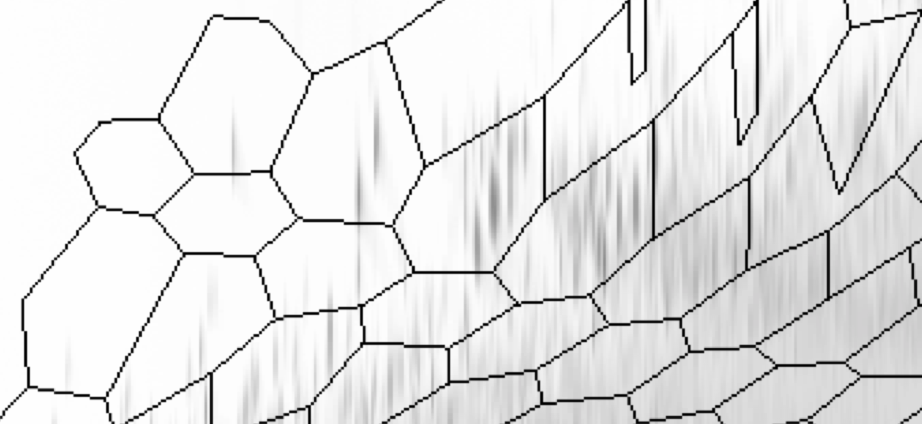}\\ 
   \includegraphics[width=0.33\textwidth, height=0.14\textwidth]{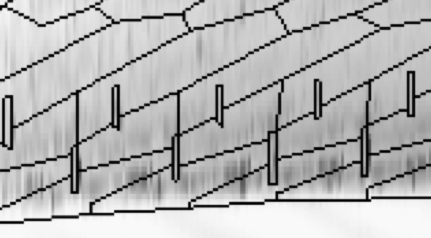}& 
   \includegraphics[width=0.33\textwidth, height=0.14\textwidth]{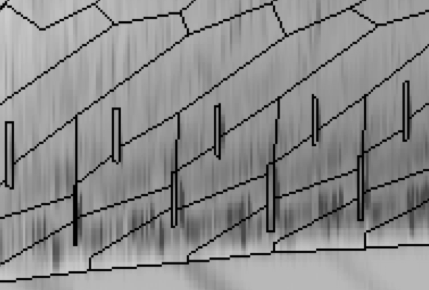}& 
   \includegraphics[width=0.33\textwidth, height=0.14\textwidth]{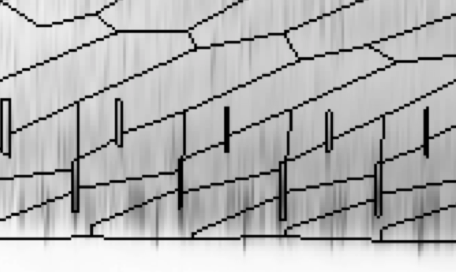}\\ 
	   Reference.  & Dataset 1  ($h = 120$). & Dataset 2 ($h = 60$).
	\end{tabular}	
\end{figure*}

The featural efficiency of the chromatogram alignment was evaluated from a more informative quantitative point of view on dataset 1. Three different ways of integrating the newly acquired chromatogram with  2DChrom\textregistered\xspace  were tested: 1) an hundred-percent manual adjustment (MA) procedure during which the user has moved every point of the template reference mask to make it match with the new data with only simple local or global translation tools; 2) the sole application of \barc alignment algorithm on the raw data and 3) the combination of \barc with light manual editing. The modified mask, after transformation \tct from flowchart in Figure \ref{fig:flowchart} with \barc, is displayed on Figure \ref{fig:reference} for both datasets 1 and 2, together with the reference template mask on the reference analysis. Concerning dataset 1, it is clearly visible that the new analysis differs from the reference analysis despite the use of the same chromatographic conditions: the new data are slightly shifted to the left and 2D retention times are higher mainly due to lower elution temperatures from the 1D column. Realignment of the template mask however looks  satisfactory, with an overall good match between the readjusted mask and the analysis. The same conclusions can be drawn for dataset 2 even if the changes between the reference chromatogram and the new one are huge as these data were not obtained with same type of modulation system. This tends to show that the algorithm is robust and is able to handle large deviations between  reference  and  new data.

Results for the quantification of chemical families are reported in Table \ref{tab:2}. These are compared with reference data previously obtained on GO-SR sample during an intra-laboratory reproducibility study on two different chromatographs with two different users so as to take into account both instrumental and user variability.

\begin{table*}
\sisetup{round-precision=1}
\centering
\caption{Dataset 1: Comparison of chemical family quantification  on  with respect to manual adjustment (MA).\label{tab:2}}
\begin{tabular}{ccccc}
 \hline
& \multicolumn{4}{c}{Weight percent}\\
  \cline{2-5}
Chemical family	& Reference 	& 100\% MA	& \barc&	\barc
$+$ MA \\
  \hline
n-\ce{C_{n}H_{2n+2}} &	\num{17.8 \pm 1.0}	  & \num{	17.2 } & \num{		12.6	 } & \num{	17.2} \\
i-\ce{C_{n}H_{2n+2}}	  & \num{	25.9 \pm 1.2	 } & \num{	25.8	 } & \num{	31.6	 } & \num{	26.0 }\\
\ce{C_{n}H_{2n}}	  & \num{	14.6 \pm 0.9 } & \num{		14.7	 } & \num{	13.9 } & \num{	14.9 }\\
\ce{C_{n}H_{2n-2}}   & \num{		11.6 \pm 0.8	 } & \num{	11.8	 } & \num{	11.7	 } & \num{	11.6 }\\
\ce{C_{n}H_{2n-6}}	  & \num{	9.5 \pm 0.7 } & \num{		9.5 } & \num{		9.4	 } & \num{	9.5 }\\
\ce{C_{n}H_{2n-8}}	  & \num{	5 \pm 0.5	 } & \num{	5.2 } & \num{		4.8 } & \num{		4.9 }\\
\ce{C_{n}H_{2n-10}}	  & \num{	2.8 \pm 0.4	 } & \num{	2.8	 } & \num{	2.8 } & \num{		2.8 }\\
\ce{C_{n}H_{2n-12}}	  & \num{	6.6 \pm 0.6 } & \num{		6.8	 } & \num{	6.6 } & \num{		6.6 }\\
\ce{C_{n}H_{2n-14}}	  & \num{	2.4 \pm 0.4 } & \num{		2.4	 } & \num{	2.4	 } & \num{	2.4 }\\
\ce{C_{n}H_{2n-16}}	  & \num{	1.6 \pm 0.3	 } & \num{	1.5 } & \num{		1.7	 } & \num{	1.7 }\\
\ce{C_{n}H_{2n-18}}	  & \num{	2 \pm 0.3 } & \num{		2.0 } & \num{		2.1	 } & \num{	2.1 }\\
\ce{C_{n}H_{2n-20}}	  & \num{	0.12 \pm 0.08 } & \num{		0.2	 } & \num{	0.2 } & \num{		0.2 }\\
\ce{C_{n}H_{2n-22}}	  & \num{	0.03 \pm 0.04	 } & \num{	0.0 } & \num{		0.1 } & \num{		0.1 }\\
  \hline
Analysis time &			& \SI{>4}{\hour} &		\SI{< 2}{\minute} &		\SI{< 30}{\minute} \\
  \hline
\end{tabular}
\end{table*}

\begin{figure}[htb]
   \caption{\label{fig:misalignment} Misalignment for thin blobs on $n$-paraffins.}
\centering
\begin{tikzpicture}
\node[anchor=south west,inner sep=0] (image) at (0,0) {\includegraphics[width=0.49\textwidth]{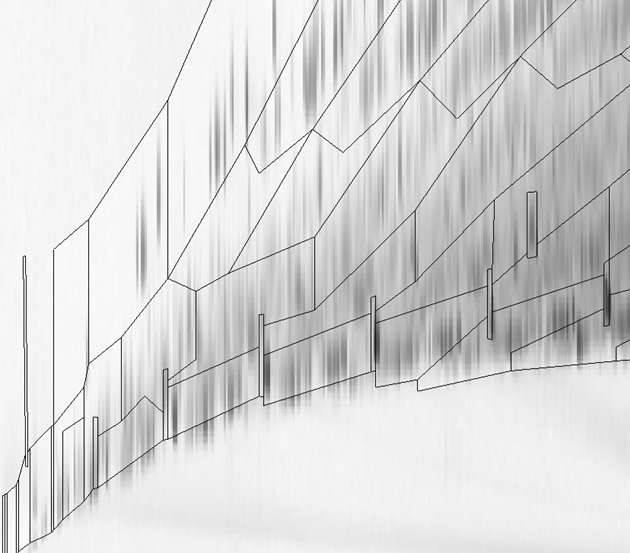}};
\begin{scope}[x={(image.south east)},y={(image.north west)}]

\draw[black,ultra thick,dashed] (0.21,0.22) circle[radius=0.4cm];
\draw[black,ultra thick,dashed,rotate around={25: (0.35,0.31)}]  (0.34,0.29) ellipse (0.06 and 0.05);
\draw[black,ultra thick,dashed,rotate around={15: (0.51,0.34)}]  (0.51,0.35) ellipse (0.07 and 0.055);
\draw[black,ultra thick,dashed,rotate around={15: (0.68,0.4)}]  (0.7,0.4) ellipse (0.07 and 0.055);

\node[](i10) at (0.21,0.12) {$i$-C10};
\node[](i11) at (0.35,0.29-0.10) {$i$-C11};
\node[](i12) at (0.50,0.34-0.09) {$i$-C12};
\node[](i13) at (0.68,0.4-0.12) {$i$-C13};

\draw[purple,ultra thick,solid] (0.275,0.235) circle[radius=0.12cm];
\draw[purple,ultra thick,solid] (0.42,0.325) circle[radius=0.12cm];
\draw[purple,ultra thick,solid] (0.6,0.375) circle[radius=0.12cm];
\draw[purple,ultra thick,solid] (0.775,0.425) circle[radius=0.12cm];

\node[purple](n10) at (0.275,0.235+0.13) {$n$-C10};
\node[purple](n11) at (0.42,0.325+0.12) {$n$-C11};
\node[purple](n12) at (0.6,0.375+0.11) {$n$-C12};
\node[purple](n13) at (0.775,0.425+0.11) {$n$-C13};
\end{scope}
\end{tikzpicture}
\end{figure}

The \barc transformation leads to coherent quantitative results for every chemical family except for normal and iso-paraffins (n-\ce{C_{n}H_{2n+2}} and i-\ce{C_{n}H_{2n+2}} respectively) and to a lesser extent naphthenes (\ce{C_{n}H_{2n}}). The quantification of $n$-paraffins is underestimated while iso-paraffins are overestimated because of slight misalignments of the template mask as depicted on the bottom of Figure \ref{fig:reference} and
on Figure \ref{fig:misalignment}. Indeed, some blobs identified in the reference mask as $n$-paraffins or naphthenes are only a few modulation periods wide as they correspond to single compounds. Small deviations in the alignment procedure  impact the accurate quantification of these blobs. An additional manual fitting is therefore required to satisfactorily correct the transformed integration mask for these specific compounds. It consists in manually moving the blob points of these small blobs to make them perfectly match with the measured individual peaks. Movements are generally smaller than one or two pixels on the first dimension and minor on the second dimension. This overall procedure is typically applied to 20 to 40 blobs for a classical gas-oil template mask and requires a few minutes.

When looking at the data analysis time required to correctly apply a sophisticated template mask on a new chromatogram, the complexity of the GO-SR sample and of the complex mask with its \num{280} blobs implies several hours of work for an experienced user with a non-automated procedure. With an anchor point based approach, at the very least  \num{50} similar points in both chromatograms would need to be defined. It would result into a processing time approaching one hour. In contrast, the processing time for dataset 1 was of two minutes, including the peak selection step. Nevertheless, depending of the samples complexity, their range of differences, and the quality of the chromatographic acquisition, the resulting masks may still require light post-processing modifications. In this case, we verified that defining typically five anchor points in an interactive registration post-processing step was enough to get a result as good as a fully manually operated one. Time saving is therefore still significant compared to manual procedures. 

\section{Conclusion}

We present in this paper a 2D-chromatogram and template alignment named \barc. It is based on three key ingredients: 1) a peak registration step which is performed on both the reference and the target 2D chromatograms; 2) two different types of transforms: a non-rigid one on the first chromatographic dimension and a rigid one on the second; 3) the use of the probabilistic  Coherent Point Drift motion estimation strategy, that is proven to be robust to noise and outliers. It results into an overall procedure that is an order of magnitude faster than the competing user-interactive alignment algorithms, with an accuracy as good as manual registration while guarantying a better reproducibility. This fast procedure may have a great interest when changing \gcgc configurations or when translating \gcgcms template masks on other \gcgc analysis (\gcgcfid analysis for example). Finally,  feature point selection may benefit from the Bayesian peak tracking recently proposed in \cite{Barcaru_A_2016_j-anal-chim-acta_bayesian_ptnpamgcgcc}.

 \section*{Acknowledgments}

The authors would like to thank Dr de Boer for his help with  \cfit. 

\bibliographystyle{unsrt}

\end{document}